\documentclass[letterpaper, 10 pt, conference]{ieeeconf} 

\IEEEoverridecommandlockouts                           
\usepackage{stfloats}
\fnbelowfloat
\usepackage{xcolor}
\usepackage{colortbl}
\usepackage{gensymb}
\usepackage{booktabs}
\usepackage{multirow}
\usepackage{graphicx}
\usepackage{amsmath}
\usepackage{amssymb}
\usepackage{cite}    
\usepackage{pdfpages}
\usepackage{balance}
\makeatletter
\let\NAT@parse\undefined
\makeatother
\usepackage{hyperref, xcolor, caption}
\usepackage{mathtools}

\newtheorem{definition}{Definition}
\usepackage{algorithm}
\usepackage{algpseudocode} 
\definecolor{lightblue}{HTML}{dfebf7}
\newcommand{\transition}{\mathcal{T}_\psi}
\algnewcommand\Input{\item[\textbf{Input:}]} 
\algnewcommand\Output{\item[\textbf{Output:}]}
\DeclareMathOperator*{\argmax}{arg\,max}

\newcommand{\ourmethod}{\textit{RoboCraft}}

\title{\LARGE \bf
Toward Humanoid Brain-Body Co-design: Joint Optimization of Control and Morphology for Fall Recovery
}
\author{Bo Yue$^{1}$, Sheng Xu$^{1}$, Kui Jia$^{1}$, Guiliang Liu$^{1}$\textsuperscript{\textdagger}
\thanks{\textdagger \ denotes the corresponding author}
\thanks{Bo Yue, Sheng Xu, Kui Jia, Guiliang Liu are with the School of Data Science, The Chinese University of Hong Kong, Shenzhen. {\tt\small boyue@link.cuhk.edu.cn, shengxu1@link.cuhk.edu.cn, kuijia@cuhk.edu.cn, liuguiliang@cuhk.edu.cn}}
}
\begin{document}

\maketitle
\thispagestyle{empty}
\pagestyle{empty}

\begin{abstract}
Humanoid robots represent a central frontier in embodied intelligence, as their anthropomorphic form enables natural deployment in humans' workspace. Brain-body co-design for humanoids presents a promising approach to realizing this potential by jointly optimizing control policies and physical morphology. 
Within this context, fall recovery emerges as a critical capability. It not only enhances safety and resilience but also integrates naturally with locomotion systems, thereby advancing the autonomy of humanoids.
In this paper, we propose \ourmethod{}, a scalable humanoid co-design framework for fall recovery that iteratively improves performance through the coupled updates of control policy and morphology.
A shared policy pretrained across multiple designs is progressively finetuned on high-performing morphologies, enabling efficient adaptation without retraining from scratch. 
Concurrently, morphology search is guided by human-inspired priors and optimization algorithms,  supported by a priority buffer that balances reevaluation of promising candidates with the exploration of novel designs. 
Experiments show that \ourmethod{} achieves an average performance gain of $44.55\%$ on seven public humanoid robots, with morphology optimization drives at least $40\%$ of improvements in co-designing four humanoid robots, underscoring the critical role of humanoid co-design.


\end{abstract}

\section{Introduction}

Can a robot be endowed with the innate potential to perform a task? Quadrupedal robots excel at stable walking and agile galloping; bipedal robots achieve upright locomotion and running; and wheel-leg hybrids combine versatility with energy-efficient traversal of uneven terrain. These bio-inspired designs demonstrate how morphology can unlock new capabilities, highlighting the foundational role of embodiment in shaping intelligent behaviors~\cite{varela2017embodied,kriegman2020scalable,whitman2023learning,badri2022birdbot}. 

Yet such choices reflect only coarse-grained body plans. The fine-grained aspects of robot design, which determine detailed structural and physical characteristics, still depend heavily on human intuition and manual intervention. Brain-body co-design aims to mitigate this reliance by jointly optimizing both the control policy and the morphological design of robotic systems \cite{Carlone2019CoDesign}, enabling robots to optimize physical structures that are inherently aligned with their expected behaviors.
Recent co-design studies have primarily targeted soft robots \cite{bhatia2021evolution,Schaff2022SoftRLDesign,Wang2023PreCo} and modular-based robots \cite{yuan2021transform2act,lu2025bodygen,jeon2025convergent}, where design spaces are less complicated and fabrication constraints are largely ignored. Humanoids, by contrast, present a far more demanding setting. Their high degrees of freedom to control, intricate body dynamics and interconnections, and strict physical constraints give rise to a vast design space with combinatorial complexity.

\begin{figure}[!t]
    \centering
    \vspace{0.1in}
    \includegraphics[page=1,width=\linewidth]{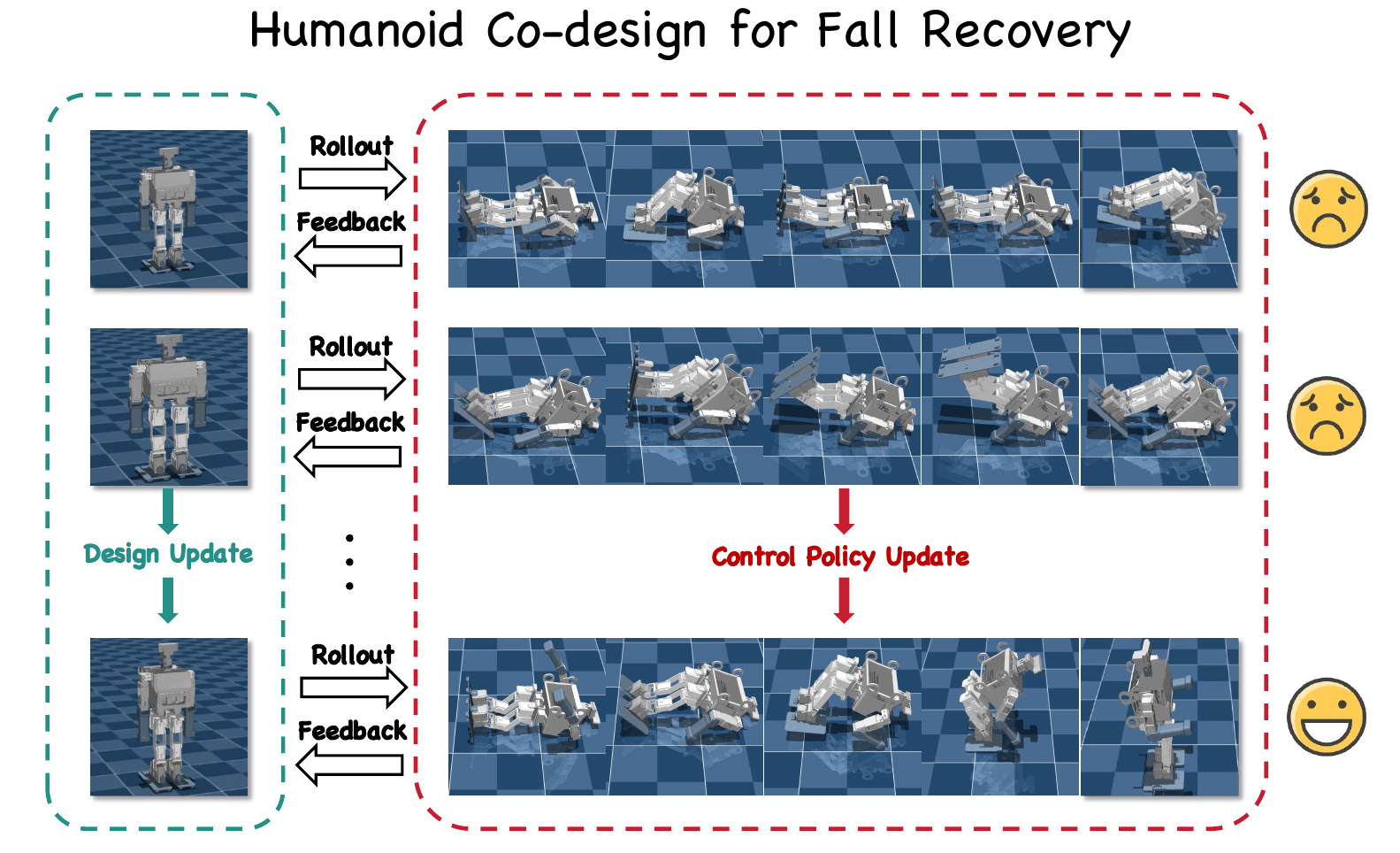}
    \caption{Co-design of morphology and control policy to enhance Bez2's recovery talent.}
    \label{fig:teaser}
\end{figure}

Within this setting, fall recovery emerges as the gateway task \cite{biomimetics9040193,stuckler2006getting,kanehiro2003first}. 
Just as infants must learn to stand before they can walk, humanoid robots must master the ability to rise after a fall before progressing to more advanced skills.
They must also reliably regain balance when inevitable disturbances occur. 
Serving both as a developmental milestone and as a safeguard for robust autonomy, fall recovery occupies a central role in humanoid co-design.
In this process, morphological optimization produces structures conducive to recovery, while control policy exploits these structures to achieve resilient performance and, in turn, provides feedback that guides further refinement of morphology. Fig. \ref{fig:teaser} illustrates an example of co-designing a humanoid robot Bez2 to enhance its recovery talent.

To advance humanoid co-design, we propose \ourmethod{}, a scalable and efficient framework that iteratively discovers improved morphologies for multiple initial designs and progressively finetunes design-specific policies derived from pretrained multi-design control policies. 
\textit{Control policy update} enhances search efficiency through both inter-design and intra-design knowledge transfer. A shared control policy is first pretrained across multiple designs, avoiding the need to retrain from scratch for each morphology. This policy is then finetuned on high-performing morphologies within each design, enabling more accurate evaluations for design-specific morphologies.  
\textit{Design update} reduces the search space through symmetry constraints, modifies the size and inertia of link STL files along with joint attributes to handle complex body dynamics, and restricts Denavit–Hartenberg parameters to maintain feasible body interconnections and physical constraints. In particular, it iteratively reevaluates promising morphologies stored in a priority buffer for more accurate assessment, samples new morphologies informed by the current control policy, and retains the top-ranking candidates in the buffer for subsequent iterations.

Experiments demonstrate that \ourmethod{} achieves an average performance improvement of $44.55\%$ on seven public humanoid robots, with morphology optimization alone driving at least $40\%$ of improvements in co-designing four humanoid robots. In addition, ablation studies of finetuning pretrained multi-design control policies prove its significance.




Our main contributions are summarized as follows.
\begin{itemize}
    \item \ourmethod{} provides a framework for humanoid co-design, accommodating multiple robots with delicate humanlike body structures and remaining compatible with various optimization algorithms.
    \item \ourmethod{} incorporates an efficient control policy update and a progressively improving design update.
    \item Experiments demonstrate that \ourmethod{}  is effective on public humanoid robots and shows the importance of taking into account the morphology optimization.
\end{itemize}

\section{Related Work}




\subsection{Brain-body Co-design}
Brain-body co-design seeks to jointly optimize a robot’s morphology and control, offering a promising pathway to promote the embodied intelligence in the dimension of physical structures. A widely adopted strategy employs bi-level optimization, where an inner reinforcement learning (RL) loop identifies effective policies for a given design, while an outer evolutionary loop searches for morphologies that yield higher-performing behaviors~\cite{geijtenbeek2013flexible,gupta2021embodied,ha2016task,bhatia2021evolution}.
Existing works have largely concentrated on bipedal~\cite{Cheng2024SERL,Ghansah2023HumanoidCoDesign}, quadrupedal~\cite{Belmonte2022RL4Leggy,Bjelonic2023CoDesignQuad,chen2024pretraining,ha2017joint,desai2018interactive}, soft~\cite{Wang2023PreCo,Li2024RL,song2025laser,Schaff2022SoftRLDesign}, and modular robots~\cite{Zhao2020RoboGrammar,Whitman2020RLDesign,lu2025bodygen,jeon2025convergent,hu2022modular,iii2021taskagnostic,Luck2019CoAdapt,pathak2019learning,wang2018neural}. However, these methods are predominantly studied in the design and simulation of soft-bodied systems or block-based robots with regular geometric shapes and simple connections, with comparatively limited progress in the practical co-design of intricate rigid-bodied robots. Extending these approaches to advanced humanoid robots with greater capabilities and physically realizable structures remains an open challenge.

\subsection{Learning Humanoid Control Policy for Fall Recovery}
For humanoid robots, fall recovery is essential for both initial standing up and regaining balance after falls. Classical stand-up control methods rely on tracking handcrafted motion trajectories through model-based optimization~\cite{kanehiro2003first,kuniyoshi2004dynamic,kanehiro2007getting,stuckler2006getting}, but these are computationally expensive and fragile to disturbances~\cite{hwangbo2019learning,lee2019robust}. More recently, RL-based methods have learned control policies with minimal modeling assumptions~\cite{peng2018deepmimic,yang2023learning,haarnoja2024learning,huang2025learning}, yielding greater robustness and real-world applicability. However, these methods remain tailored to individual robots and fail to generalize across embodiments. 
While some approaches explicitly encode morphology into policy learning and robot design~\cite{sferrazza2024body,yuan2021transform2act,lu2025bodygen,bohlinger2024one,huang2020one,xiong2023universal}, unified policies have been developed that exploit morphology implicitly from observations to generalize fall recovery across diverse humanoid robots without structural priors~\cite{spraggett2025unihum,frasa2024}. Nevertheless, these efforts remain control-centric and overlook the potential to optimize morphology for fall recovery. As a result, they offer limited guidance to humanoid design, leaving unexplored the opportunity to shape morphologies that are inherently more capable of recovering from falls.

\section{Problem Formulation}
\textbf{Humanoid Robot Learning Environment.} To specify the learning environment for humanoid robot co-design, conditioning on a specific robot design $\psi$ from the design space $\Psi$, we considered a Markov Decision Process (MDP) $\mathcal{M}(\psi)=(\mathcal{S}, \mathcal{A}, \transition, r,  \mu_0^\psi, \gamma)$, where: 
\begin{itemize}
 \item The state $s_t\in\mathcal{S}$ captures the robot's posture, motion, and recent control history.
 \item The action $a_t\in\mathcal{A}$ encodes angular-velocity increments for a fixed set of joints. These increments are integrated to yield target joint angles, which a PD controller then tracks to actuate the corresponding degrees of freedom (DoFs).
 \item The transition function $\transition$ is a mapping from the state-action pairs to a probability distribution over the future state space, i.e., $\transition: \mathcal{S} \times \mathcal{A} \to \Delta(\mathcal{S})$.
 \item $r(\cdot):\mathcal{S}\times\mathcal{A}\to\mathcal{R}$ represents the reward function that encourages the robot to accomplish a task.
 \item $\mu_0^\psi\in\Delta({\mathcal{S}})$ denotes the initial state distribution.
 \item $\gamma\in[0,1)$ is the discount factor.
\end{itemize}

We assume $(\mathcal{S},\mathcal{A},r)$ fixed across all $\psi$, while allowing the dynamics and initial distribution to vary with $\psi$, i.e., $(\mathcal{T}_\psi,\mu_0^\psi)$. This assumption is reasonable for humanoids, as they typically share state and action spaces with consistent semantic meaning across dimensions. Moreover, the reward functions are designed to be morphology-agnostic, capturing physical performance in a manner that is independent of the specific body design.

\textbf{Bi-Level Optimization for Humanoid Robot Co-design}.
The robot co-design problem involves the joint design of both software components (e.g., control policies) and hardware components (e.g., robotic modules) to optimize task-dependent performance~\cite{Carlone2019CoDesign}. When generalizing this problem to a learning-based formulation, the bi-level optimization framework commonly involves a forward pass for policy training and a backward pass for updating the robot design.

Specifically, during the forward process, we use the RL algorithm to optimize the policy function parameterized by $\theta$, i.e., learning $\pi_\theta(\psi):\mathcal{S}\to\Delta(\mathcal{A})$, to maximize the expected return of the discounted cumulative rewards:
\begin{align}
    \max_{\theta\in \Theta}\mathcal{J}(\pi_\theta, \mathcal{M}(\psi)) = \max_{\theta\in \Theta}\mathbb{E}_{ \pi_\theta,\transition,\mu_0^\psi}\left[\sum_{t=0}^{\infty} \gamma^t r(s_t,a_t)\right]
\end{align}
where the expectation is taken over the policy, the transition dynamics and initial state distribution, conditioning on the MDP associated with the given robot design $\mathcal{M}(\psi)$.

In addition to updating the policy, the inverse update on the robot's structure necessitates formulating humanoid robot co-design as a bi-level optimization problem:
\begin{align}
    \max_{\psi\in\Psi} \max_{\theta\in \Theta} \mathcal{J}(\pi_\theta, \mathcal{M}(\psi)) \label{obj:bi-level}
\end{align}
Note that there is no guarantee that such a max-max problem is well-defined, so it often requires additional conditions, which we present as follows. 

\begin{figure*}[!t]
    \centering
    \includegraphics[page=2,width=\linewidth]{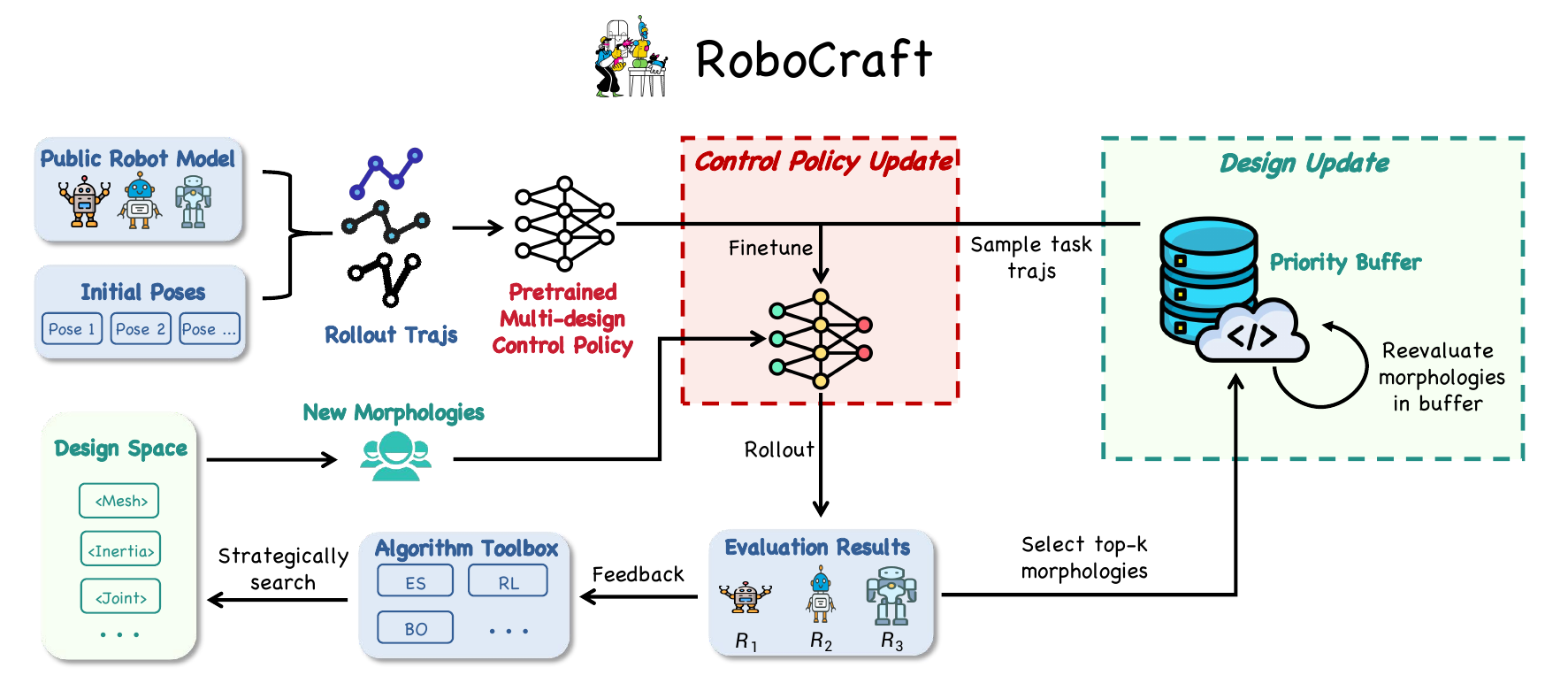}
    \caption{The \ourmethod{} framework.}
    \label{fig:framework}
\end{figure*}

\begin{definition} \textit{(Conditions of Learning-based Co-design Problem).} The co-design problem~(\ref{obj:bi-level}) should satisfy the following conditions:
\begin{enumerate}
    \item \textit{(Bounded rewards)}: The reward function in \( \mathcal{M}(\psi)\) is bounded by $r(\cdot)\in[0, R_{\max}]$.
    \item \textit{(Compact Feasible Sets)}: The policy parameterization space and the design space, $\Theta$ and $\Psi$, are compact.
    \item \textit{(Continuity)}: \(\mathcal{J}(\pi_\theta, \mathcal{M}(\psi))\) is continuous in \(\Theta\) and \(\Psi\).
\end{enumerate}

\end{definition}
Under these conditions, by the Extreme Value Theorem (EVT), there exists $(\theta^\star,\psi^\star)\in\Theta\times\Psi$ attaining the maximum.

\section{Method\label{sec:method}}
To address this bi-level optimization problem, we propose \ourmethod{}, a scalable and efficient humanoid co-design framework that progressively uncovers novel, ever-advancing morphologies guided by finetuned policies derived from pretrained multi-design control policies.

In the forward phase of \textit{control policy update}, we first pretrain a shared control policy across several initial designs of humanoid robots. This multi-design policy serves as a coarse task evaluator of morphology performance. Based on its evaluations, the design update phase iteratively explores new morphologies for each design, collects their task trajectories, and uses them to finetune the shared control policy, thereby producing a stronger evaluator. 
In the backward phase of \textit{design update}, algorithms such as evolutionary search or Bayesian optimization strategically explore new morphologies of each design guided by the evaluator's feedback. At each iteration, top-performing morphologies from previous iterations are reevaluated and compared with newly-generated morphologies by the newest control policy. This ensures that only the most promising morphologies are retained for the subsequent iteration.
We detail control policy update in Sec. \ref{sec:forward}, design update in Sec. \ref{sec:backward}, and the \ourmethod{} framework in Sec. \ref{sec:framework}, with Fig.~\ref{fig:framework} illustrating the overall framework.

\subsection{Forward Phase: Control Policy Update}\label{sec:forward}
To assess the performance of a morphology evolved from an initial design, prior co-design approaches typically learn a task policy from scratch, without leveraging shared experience either across designs or among the evolving morphologies within each design. 
We realize a two-fold knowledge transfer to address the limitations: (i) \textit{inter-design transfer}, by pretraining a shared control policy across multiple designs to avoid retraining from scratch; and (ii) \textit{intra-design transfer}, by finetuning this policy on the top-ranking morphologies within each design to capture morphology-specific adaptations to mitigate redundant learning.

In inter-design transfer, we follow \cite{spraggett2025unihum,frasa2024} and pretrain a shared control policy $\pi_{\theta_0}$ (with $0$ denoting the initial co-design iteration) by collecting trajectories from several initial humanoid designs in open-source robot models. 
The initial poses of the humanoid robots in each episode are randomized to emulate arbitrary fall configurations.
Although coarse, $\pi_{\theta_0}$ offers a lower bound on task performance for $n$-th initial design $\psi^0_n$:
\begin{align}
    \mathcal{J}(\pi_{\theta^*}, \mathcal{M}(\psi^0_n)) 
    \geq\mathcal{J}(\pi_{\theta_0}, \mathcal{M}(\psi^0_n)), \label{obj:lower-bound}
\end{align}
where the optimal policy parameters for the design,
\begin{align}
    \theta^* =\argmax\limits_{\theta} \mathcal{J}(\pi_{\theta}, \mathcal{M}(\psi^0_n)).
\end{align}

In intra-design transfer, we propose to collect trajectories from the top-ranking morphologies within each design and use them to finetune the control policy from the previous iteration. 
This recursive process yields progressively more specialized control policies, thereby producing evaluations of intra-design morphologies that increasingly approximate the optimal. For a morphology $\psi^{i-1}_n$ at co-design iteration $i-1(i\geq 1)$, control policy update is formalized as:
\begin{align}
    \mathcal{J}(\textcolor{purple}{\pi_{\theta_{i}}}, \mathcal{M}(\psi^{i-1}_n)) 
    \geq\mathcal{J}(\textcolor{purple}{\pi_{\theta_{i-1}}}, \mathcal{M}(\psi^{i-1}_n)). \label{obj:recursive-improvement-policy}
\end{align}

Both forms of knowledge transfer substantially improve the sample efficiency of control policy learning and provide effective feedback to the design update phase.

\subsection{Backward Phase: Design Update}\label{sec:backward}
Although co-design has been applied to other robotic forms, its extension to advanced humanoids with carefully engineered, physically feasible structures is still largely uncharted.

Tab.~\ref{tab:xml_modifications_implemented} defines the design search space. In the MuJoCo simulator~\cite{todorov2012mujoco}, this space is realized through XML element modifications that adjust the mesh dimensions of link STL files, the physical properties of links, and the dynamic attributes of joints. The initial designs are specified in MJCF (MuJoCo XML format), converted from URDF (Unified Robot Description Format) files of publicly available humanoid robots.

\begin{table}[!t]
\centering
\caption{Design space parameters for humanoid morphology modification.}
\label{tab:xml_modifications_implemented}
\renewcommand\arraystretch{1.2} 
\resizebox{0.5\textwidth}{!}{%
\begin{tabular}{l p{4cm} p{4cm}}
\toprule
\textbf{Category} & \textbf{Meaning} & \textbf{Modified XML Elements} \\
\midrule
Mesh Scaling & Scales meshes in X-dimension & \texttt{<asset><mesh>} \\
             & Scales meshes in Y-dimension & \texttt{<mesh>} \\
             & Scales meshes in Z-dimension & \texttt{<mesh>} \\ \hline
Mass Scaling & Scales masses                 & \texttt{<inertial>} \\ \hline
Joint Stiffness Scaling & Scales joint stiffness & \texttt{<stiffness>} \\ \hline
Joint Damping Scaling   & Scales joint damping   & \texttt{<damping>} \\
\bottomrule
\end{tabular}%
}
\end{table}

Our framework is compatible with a broad class of optimization algorithms for searching the next morphology based on previous evaluations (feedback). In particular, we deliver an algorithm toolbox that incorporates evolutionary search, the covariance matrix adaptation evolution strategy (CMA-ES), Bayesian optimization, and RL.
At iteration $0$, for every initial design $n$, we sample multiple rounds of morphologies, evaluate each using the pretrained control policy $\pi_{\theta_0}$, and store the top-$k$ performers in a priority buffer $B^0_n$, where $k$ is a hyperparameter. At each subsequent iteration $i \geq 1$, the morphologies in $B^i_n$ are reevaluated under the current control policy $\pi_{\theta_i}(\psi_n^i)$ and compared with newly sampled morphologies $S^i_n$. The buffer is then updated according to
\begin{align}
    B^{i}_n = \text{Top-}k\big(B^{i-1}_n \cup S^i_n\big),
\end{align}
where $\text{Top-}k(\cdot)$ denotes the operator that retains the $k$ highest-performing morphologies. Suppose the buffer replace $\psi^{i-1}_n$ with $\psi^{i}_n$; design update is formalized as:
\begin{align}
    \mathcal{J}(\pi_{\theta_i}, \mathcal{M}(\textcolor{teal}{\psi^{i}_n}))
    \geq\mathcal{J}(\pi_{\theta_i}, \mathcal{M}(\textcolor{teal}{\psi^{i-1}_n})). \label{obj:recursive-improvement-morph}
\end{align}

\begin{algorithm}[!ht]
\caption{Iterative Joint Optimization of Control Policy and Morphology in Humanoid Robots}\label{alg:main}
\begin{algorithmic}[1]
\Input training iterations $N_i$, sample rounds $N_r$, num. of samples $N_s$, num. of initial designs $N_d$, priority buffer $B^{0}_n$, buffer size $K$, optimization algorithm $\Omega$
\Output control policy, buffer of morphologies
\State Initialize the control policy as $\pi_{\theta_0}$
\State Sample trajectories from $N_d$ initial humanoid designs
\State Pretrain $\pi_{\theta_0}$ with sampled trajectories
\For{$n \gets 0 \text{ to } N_d-1$}
\For{$i \gets 0 \text{ to } N_i-1$}
    \For{$j \gets 0 \text{ to } N_r-1$}
        \State Strategically search $N_s$ morphologies with $\Omega$
        \State Evaluate each morphology with $\pi_{\theta_i}$ 
    \EndFor
    \If{$i=0$}
        \State Select top-$K$ morphologies from samples
        \State Initialize buffer $B^{i}_n$ with these morphologies
    \Else
        \State Reevaluate the morphologies in buffer $B^{i-1}_n$ with $\pi^i_\theta$ 
        \State Select top-$K$ morphologies from the union of buffer $B^{i-1}_n$ and samples in current iteration
        \State Push selected morphologies into buffer $B^{i}_n$
        \State Retain top-$K$ morphologies in buffer $B^{i}_n$ and abandon the remaining \textcolor{teal}{\Comment{Design update}}
    \EndIf
    \State Sample trajectories from morphologies in $B^{i}_n$
    \State Finetune $\pi^i_\theta$ into $\pi^{i+1}_\theta$ \textcolor{purple}{\Comment{Control policy update}}
\EndFor
\EndFor
\State \Return $\pi^{N_i}_{\theta_i}$ and $B^{N_i}_n$ for each design
\end{algorithmic}
\end{algorithm}

\subsection{\ourmethod{} Framework}\label{sec:framework}
By combining the control policy update (Eq. (\ref{obj:recursive-improvement-policy})) and the design update (Eq. (\ref{obj:recursive-improvement-morph})), we derive the iterative bi-update process as:
\begin{align}
    \mathcal{J}(\textcolor{purple}{\pi_{\theta_i}}, \mathcal{M}(\textcolor{teal}{\psi^i_n}))
    \geq\mathcal{J}(\textcolor{purple}{\pi_{\theta_{i-1}}}, \mathcal{M}(\textcolor{teal}{\psi^{i-1}_n})). \label{obj:recursive-improvement-combine}
\end{align}
This bi-update procedure is repeated for $N_i$ iterations for each initial design, producing a specialized control policy and its top-$k$ evolved morphologies. The overall algorithm is summarized in Alg.~\ref{alg:main}.

\section{Experiments}
We report our experimental setup in Sec. \ref{sec:expa}, and organize our experiments to answer the following three research questions (RQs):
\begin{itemize}
    \item \textit{RQ1 (Effectiveness of \ourmethod{}).} How effective is \ourmethod{} in co-designing humanoid robots for fall recovery? (Sec. \ref{sec:exp1})
    \item \textit{RQ2 (Role of control policy vs. morphology).} What are the respective contributions of control policy and morphology optimization to elevating the performance of humanoid fall recovery? (Sec. \ref{sec:exp2})
    \item \textit{RQ3 (Necessity of control policy fine-tuning).} Can we still obtain high-performing morphologies without fine-tuning a pre-trained control policy? (Sec. \ref{sec:exp3})
\end{itemize}

\begin{table*}[!tb]
\caption{Fall recovery performance of seven humanoid robots optimized through co-design.}
\label{tab:effectiveness}
\centering
\renewcommand\arraystretch{1.1} 
\resizebox{0.8\textwidth}{!}{
\begin{tabular}{cccccccc}
\toprule
\multirow{2}{*}{\textbf{Methods}} & \multicolumn{7}{c}{\textbf{Humanoid Robots}} \\ \cline{2-8} 
 & \textbf{Bez1} & \textbf{Bez2} & \textbf{Bez3} & \textbf{OP3-Rot} & \textbf{Sigmaban} & \textbf{Wolfgang} & \textbf{NUGUS} \\ \hline
Base (initial design) & 27.36 & 81.79 & 100.59 & 64.53 & 52.29 & 134.88 & 107.28 \\ 
\rowcolor{lightblue} RL & 59.31 & 81.66 & \textbf{120.36} & 81.41 & 102.91 & 88.04 & 105.29 \\ 
\rowcolor{lightblue} Evolutionary search & \textbf{66.89} & \textbf{97.98} & 119.63 & 82.36 & \textbf{107.92} & \textbf{109.84} & 93.14 \\ 
\rowcolor{lightblue} Bayesian optimization & 64.06 & 93.44 & 119.00 & \textbf{85.11} & 101.43 & 102.43 & 108.18 \\ 
\rowcolor{lightblue} CMA-ES & 62.09 & 88.00 & 118.05 & 79.52 & 105.87 & 91.31 & \textbf{116.90} \\ 
\bottomrule
\end{tabular}
}
\end{table*}

\subsection{Experimental Setup}\label{sec:expa}

In control policy update, we follow the setup of \cite{spraggett2025unihum}. We co-design on seven open-source humanoid robots (Bez1, Bez2, Bez3, OP3-Rot, Sigmaban, Wolfgang, and NUGUS), adopting their unified control setup and applying control at 20 Hz for smooth, stable operation.
The observation space is a $27$-dimensional vector comprising joint positions $q_t$, velocities $\dot{q_t}$, and desired targets $q_t^{\text{desired}}$ ($5$ each) for five sagittal joints (shoulder pitch, elbow, hip pitch, knee, ankle), torso orientation $\theta_t^{r,p,y}$ and angular velocity $\dot{\theta}_t^{r,p,y}$ ($3$ each), normalized head height $h^{\text{head}}_t$ ($1$), and the previous action $a_{t-1}$ ($5$). No morphology identifiers are included.
The action space is morphology-independent and outputs velocity increments, applied symmetrically on both sides, $a_t=\dot{q}_t^{\text{desired}}$ ($5$). These increments are integrated into target joint angles at each $50$~ms step, with MuJoCo enforcing joint limits automatically.
The morphology-agnostic reward function encourages upright recovery while promoting smooth and safe movements. It is defined as the unweighted sum of five terms: an upright reward $R_{\text{Up}}=\exp(-10\|h^{\text{head}}_t-1.0\|^2)$ encourages head height; a torso alignment reward $R_{\text{Pitch}}=\mathbf{1}_{\{h^{\text{head}}_t>0.4\}}\exp(-10\|\theta^{p}_t\|^2)$ activates once the robot has partially risen; and three small penalties regularize behavior, $R_{\text{vel}}=0.1\exp(-\|\dot{q}_t\|)$ for joint velocity, $R_{\text{var}}=0.05\exp(-\|a_t-a_{t-1}\|)$ for abrupt action changes, and $R_{\text{collision}}=0.1\exp(-\text{selfCollision}_t)$ for self-collisions. 

The initial poses of the humanoid in each episode are randomized to simulate arbitrary fall, and a domain randomization scheme \cite{tan2018sim} is applied to model unexpected errors as in \cite{frasa2024}.
We pretrain the shared policy on $7$ initial designs, using Soft Actor–Critic (SAC)\cite{haarnoja2018soft}, augmented with the CrossQ algorithm\cite{raffin2021stable}. Pretraining is conducted for $6\times10^5$ timesteps with an MLP policy, a learning rate of $1\times10^{-3}$, and a discount factor $\gamma=0.99$. The policy network follows a three-layer architecture with hidden sizes $[512, 512, 256]$. Subsequent finetuning is conducted for $1\times10^5$ timesteps for trajectories collected from each selected morphology.

In design update, we run 100 iterations, each consisting of sampling 3 rounds of 10 morphologies from the design space. The size of the priority buffer is $5$. We assign the name to each unique XML by computing a hash over its contents, ensuring every morphology file has a consistent and conflict-free identifier. For every sampled morphology, we evaluate its episode rewards $10$ times and take the average. Each episode runs for up to $10s$ of simulation time and is terminated early for safety if the torso tilts more than $135\degree$ from upright or if the angular velocity exceeds $25\degree/s$.
In MJCF models, the same mesh geoms serve simultaneously as the visual representation and the collision geometry, such that the rendered surfaces directly correspond to those used for contact simulation in MuJoCo.

Morphology optimizers operate over bounded continuous parameter spaces, ensuring that neighboring links remain interconnected as in the initial design and that morphologies respect physical limits in MuJoCo.
Four morphology optimizers were employed with specific configurations. The RL optimizer uses a policy gradient method with a two-layer MLP ($64$ units each), a learning rate of $1\times10^{-3}$, an exploration noise of $0.1$, and an experience replay buffer (size $1000$, batch size $32$). The evolutionary search algorithm uses tournament selection (tournament size $3$), blend crossover (BLX-$\alpha$ with $\alpha=0.3$), and Gaussian mutation (mutation rate $0.1$, strength $0.1$), with a population size of $50$ and $10\%$ elitism. The Bayesian optimization method employs a Gaussian Process with an RBF kernel (length scale $1.0$, noise level $1\times10^{-5}$) and an Expected Improvement acquisition function, after an initial design of $10$ random samples to seed the model. Finally, the CMA-ES optimizer starts from the center of the parameter space with an initial step size $0.3$ and adaptively updates its covariance matrix (as well as the step size) over successive generations. Each optimization algorithm requires about $30$ hours of computation on an AMD EPYC 7763 64-Core CPU paired with an NVIDIA RTX 4090 GPU.

\subsection{Effectiveness of \ourmethod{}}\label{sec:exp1}
To empirically validate \ourmethod{}, we evaluate the co-design performance of final-iteration morphologies together under finetuned control policies. Tab.~\ref{tab:effectiveness} summarizes the resulting episode rewards across seven humanoid robots on four optimizers. The base method reports the episode rewards of seven initial designs on the pretrained control policy. 
Experiments demonstrate that \ourmethod{} improves task performance by an average of $44.55\%$.
Except for robot Wolfgang, the best optimizers (bold) achieve higher rewards than the baseline and are particularly effective for initial designs with poor performance, such as Bez1 ($27.26\to66.89, +145\%$), OP3-Rot ($64.53\to85.11, +32\%$), and Sigmaban ($52.29\to107.92, +106\%$). 
Since Wolfgang is already capable of standing up with its initial design, co-design provides little additional benefit. Among the methods, evolutionary search achieves the best performance in co-designing $4$ out of $7$ robots.
\begin{figure}[!t]
    \centering
    \includegraphics[page=3,width=0.8\linewidth]{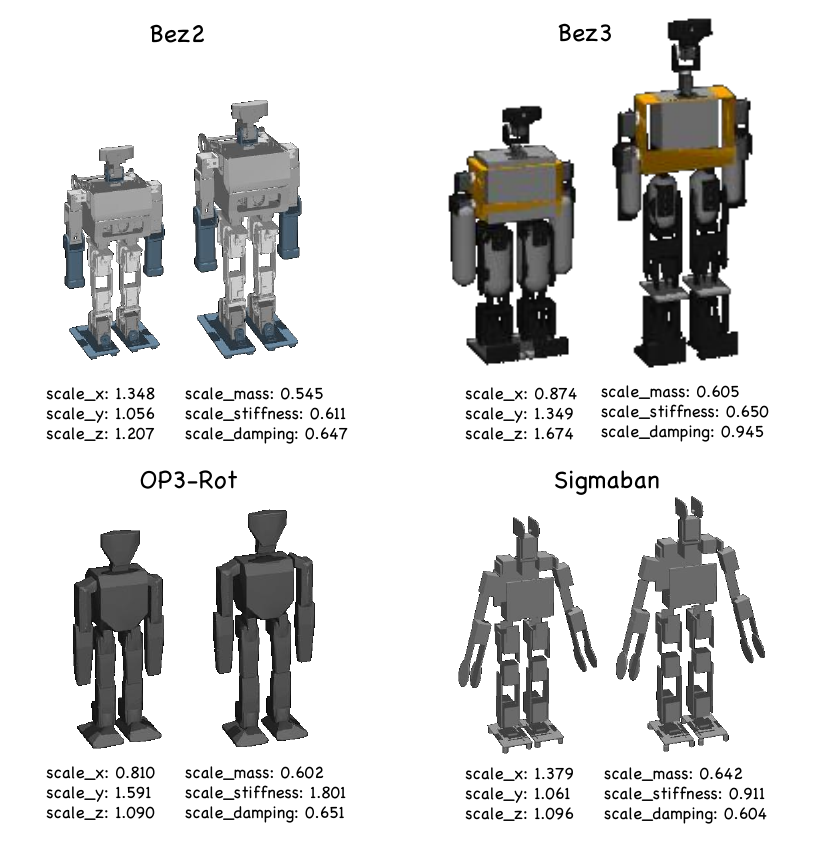}
    \caption{Initial designs (left) vs. optimized morphologies (right) of Bez2, Bez3, OP3-Rot and Sigmaban.}
    \label{fig:effectiveness}
\end{figure}

We also present in Fig.~\ref{fig:effectiveness} the initial designs (left) and the optimized morphologies (right) of Bez2, Bez3, OP3-Rot, and Sigmaban as case studies. We find that taller robots are evolved due to the upright reward term in the task reward function. In addition, lowering passive resistance makes it easier for the controller to generate the torques needed to stand up from the ground.

\subsection{Role of Control Policy vs. Morphology}\label{sec:exp2}
Co-designing a humanoid robot enhances task performance by jointly optimizing the control policy and the morphology. To assess their respective contributions, we quantify their individual effects within the co-design process, which is essential for understanding whether the observed improvements arise from an evenly matched synergy or are dominated by a single factor. Such analysis clarifies when co-design is necessary and when single-focus optimization might suffice.
We decompose the improvement of the bi-level optimization objective, $\Delta\mathcal{J}_n$, measured from the $n$-th initial design under a pretrained control policy to the optimized design under a fine-tuned policy at the final iteration, into two components: Eq.~(\ref{eq:2-1-general}) captures the contribution of control policy optimization, and Eq.~(\ref{eq:2-2-general}) captures the contribution of morphology optimization. This decomposition follows directly from splitting Eq. (\ref{eq:half-general}) into two symmetric halves in step ($a1$) and then permuting the terms from the order (1,2,3,4) to (1,4,3,2) in step ($b1$). Here, $d$ is the step size or the critical point to divide two kinds of contributions in Eqs. (\ref{eq:d-1}) and (\ref{eq:d-2}). We use $d=N_i$ for our experiments, and the two contribution components are Eqs.~(\ref{eq:2-1}) and (\ref{eq:2-2}).
\begin{figure}[!tb]
    \centering
    \includegraphics[page=1,width=\linewidth]{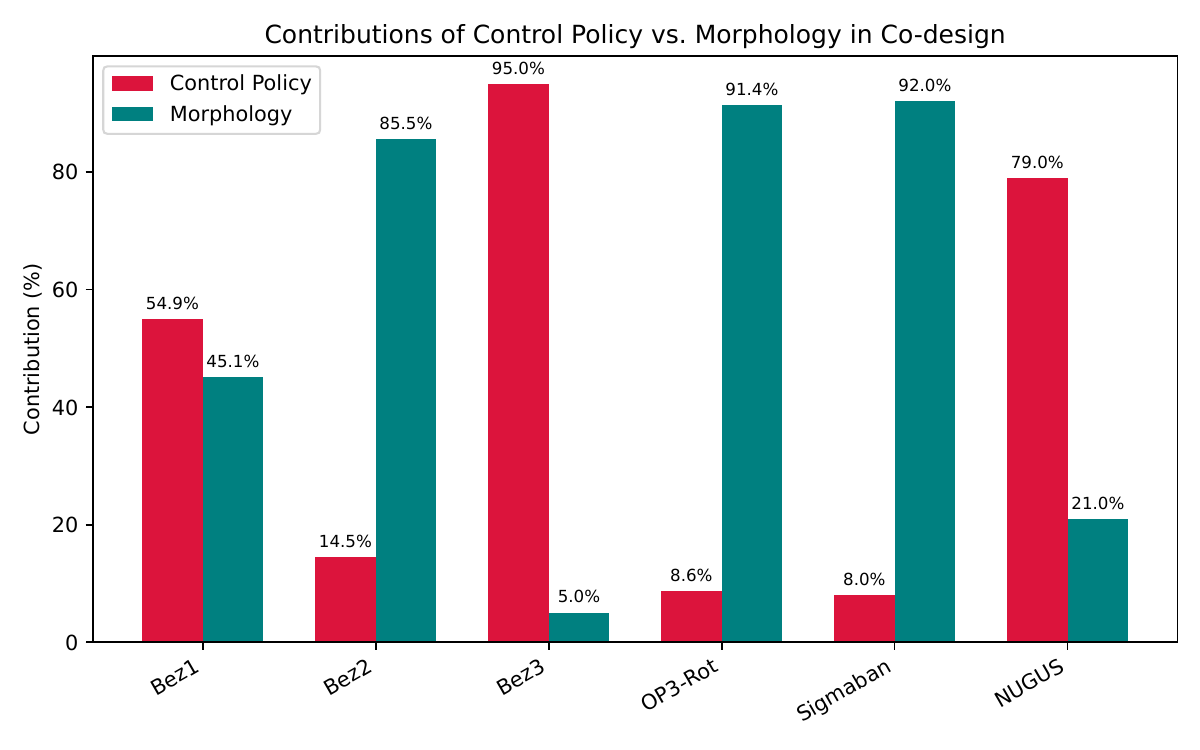}
    \caption{Contribution of control policy vs. morphology optimization in co-design.}
    \label{fig:contribution}
\end{figure}

\begin{figure*}[!tb]
\normalsize
\begin{align}
    \Delta\mathcal{J}_n\;=\;\mathcal{J}(&\textcolor{purple}{\pi_{\theta_{N_i}}}, \mathcal{M}(\textcolor{teal}{\psi^{N_i}_n}))
    -\mathcal{J}(\textcolor{purple}{\pi_{\theta_{0}}}, \mathcal{M}(\textcolor{teal}{\psi^{0}_n}))\label{eq:half-general}\\
    \overset{(a1)}{=}\;\frac{1}{2}\Bigg[&\sum_{\lambda=\frac{N_i}{d}}^{1}\Big[\underbrace{\big[\mathcal{J}(\textcolor{purple}{\pi_{\theta_{\lambda d}}}, \mathcal{M}(\textcolor{teal}{\psi^{\lambda d}_n}))
    -\mathcal{J}(\textcolor{purple}{\pi_{\theta_{(\lambda-1) d}}}, \mathcal{M}(\textcolor{teal}{\psi^{\lambda d}_n}))
    \big]}_{\text{contribution of \textit{control policy} optimization term 1}}+\underbrace{\big[\mathcal{J}(\textcolor{purple}{\pi_{\theta_{(\lambda-1) d}}}, \mathcal{M}(\textcolor{teal}{\psi^{\lambda d}_n}))-\mathcal{J}(\textcolor{purple}{\pi_{\theta_{(\lambda-1) d}}}, \mathcal{M}(\textcolor{teal}{\psi^{(\lambda-1) d}_n}))\big]\Big]}_{\text{contribution of \textit{morphology} optimization term 2}}\Bigg]\label{eq:d-1}\\
    +\; \frac{1}{2}\Bigg[&\sum_{\lambda=\frac{N_i}{d}}^{1}\Big[\underbrace{\big[\mathcal{J}(\textcolor{purple}{\pi_{\theta_{\lambda d}}}, \mathcal{M}(\textcolor{teal}{\psi^{\lambda d}_n}))
    -\mathcal{J}(\textcolor{purple}{\pi_{\theta_{\lambda d}}}, \mathcal{M}(\textcolor{teal}{\psi^{(\lambda-1) d}_n}))
    \big]}_{\text{contribution of \textit{morphology} optimization term 1}}+\underbrace{\big[\mathcal{J}(\textcolor{purple}{\pi_{\theta_{\lambda d}}}, \mathcal{M}(\textcolor{teal}{\psi^{(\lambda-1) d}_n}))-\mathcal{J}(\textcolor{purple}{\pi_{\theta_{(\lambda-1) d}}}, \mathcal{M}(\textcolor{teal}{\psi^{(\lambda-1) d}_n}))\big]\Big]}_{\text{contribution of \textit{control policy} optimization term 2}}\Bigg] \label{eq:d-2}\\
    \overset{(b1)}{=}\;\frac{1}{2}\Bigg[&\sum_{\lambda=\frac{N_i}{d}}^{1}\Big[\underbrace{\big[\mathcal{J}(\textcolor{purple}{\pi_{\theta_{\lambda d}}}, \mathcal{M}(\textcolor{teal}{\psi^{\lambda d}_n}))
    -\mathcal{J}(\textcolor{purple}{\pi_{\theta_{(\lambda-1) d}}}, \mathcal{M}(\textcolor{teal}{\psi^{\lambda d}_n}))
    \big]}_{\text{contribution of \textit{control policy} optimization term 1}}+\underbrace{\big[\mathcal{J}(\textcolor{purple}{\pi_{\theta_{\lambda d}}}, \mathcal{M}(\textcolor{teal}{\psi^{(\lambda-1) d}_n}))-\mathcal{J}(\textcolor{purple}{\pi_{\theta_{(\lambda-1) d}}}, \mathcal{M}(\textcolor{teal}{\psi^{(\lambda-1) d}_n}))\big]\Big]}_{\text{contribution of \textit{control policy} optimization term 2}}\Bigg]\label{eq:2-1-general}\\
    +\; \frac{1}{2}\Bigg[&\sum_{\lambda=\frac{N_i}{d}}^{1}\Big[\underbrace{\big[\mathcal{J}(\textcolor{purple}{\pi_{\theta_{\lambda d}}}, \mathcal{M}(\textcolor{teal}{\psi^{\lambda d}_n}))
    -\mathcal{J}(\textcolor{purple}{\pi_{\theta_{\lambda d}}}, \mathcal{M}(\textcolor{teal}{\psi^{(\lambda-1) d}_n}))
    \big]}_{\text{contribution of \textit{morphology} optimization term 1}}+\underbrace{\big[\mathcal{J}(\textcolor{purple}{\pi_{\theta_{(\lambda-1) d}}}, \mathcal{M}(\textcolor{teal}{\psi^{\lambda d}_n}))-\mathcal{J}(\textcolor{purple}{\pi_{\theta_{(\lambda-1) d}}}, \mathcal{M}(\textcolor{teal}{\psi^{(\lambda-1) d}_n}))\big]\Big]}_{\text{contribution of \textit{morphology} optimization term 2}}\Bigg]\label{eq:2-2-general}
\end{align}
\end{figure*}

\begin{figure*}[!tb]
\normalsize
\begin{align}
    \Delta\mathcal{J}_n\;=\;\mathcal{J}(&\textcolor{purple}{\pi_{\theta_{N_i}}}, \mathcal{M}(\textcolor{teal}{\psi^{N_i}_n}))
    -\mathcal{J}(\textcolor{purple}{\pi_{\theta_{0}}}, \mathcal{M}(\textcolor{teal}{\psi^{0}_n})) \label{eq:half}\\
    \overset{(a2)}{=}\;\frac{1}{2}\Bigg[&\Big[\mathcal{J}(\textcolor{purple}{\pi_{\theta_{N_i}}}, \mathcal{M}(\textcolor{teal}{\psi^{N_i}_n}))-\mathcal{J}(\textcolor{purple}{\pi_{\theta_{0}}}, \mathcal{M}(\textcolor{teal}{\psi^{N_i}_n}))\Big]
    +\Big[\mathcal{J}(\textcolor{purple}{\pi_{\theta_{0}}}, \mathcal{M}(\textcolor{teal}{\psi^{N_i}_n}))-\mathcal{J}(\textcolor{purple}{\pi_{\theta_{0}}}, \mathcal{M}(\textcolor{teal}{\psi^{0}_n}))\Big]\Bigg] \nonumber\\
    +\;\frac{1}{2}\Bigg[&\Big[\mathcal{J}(\textcolor{purple}{\pi_{\theta_{N_i}}}, \mathcal{M}(\textcolor{teal}{\psi^{N_i}_n}))-\mathcal{J}(\textcolor{purple}{\pi_{\theta_{N_i}}}, \mathcal{M}(\textcolor{teal}{\psi^{0}_n}))\Big]
    +\Big[\mathcal{J}(\textcolor{purple}{\pi_{\theta_{N_i}}}, \mathcal{M}(\textcolor{teal}{\psi^{0}_n}))-\mathcal{J}(\textcolor{purple}{\pi_{\theta_{0}}}, \mathcal{M}(\textcolor{teal}{\psi^{0}_n}))\Big]\Bigg]\nonumber\\
    \overset{(b2)}{=}\;\frac{1}{2}\Bigg[&\Big[\mathcal{J}(\textcolor{purple}{\pi_{\theta_{N_i}}}, \mathcal{M}(\textcolor{teal}{\psi^{N_i}_n}))-\mathcal{J}(\textcolor{purple}{\pi_{\theta_{0}}}, \mathcal{M}(\textcolor{teal}{\psi^{N_i}_n}))\Big]
    +\Big[\mathcal{J}(\textcolor{purple}{\pi_{\theta_{N_i}}}, \mathcal{M}(\textcolor{teal}{\psi^{0}_n}))-\mathcal{J}(\textcolor{purple}{\pi_{\theta_{0}}}, \mathcal{M}(\textcolor{teal}{\psi^{0}_n}))\Big]\Bigg]\label{eq:2-1}\\
    +\;\frac{1}{2}\Bigg[&\Big[\mathcal{J}(\textcolor{purple}{\pi_{\theta_{N_i}}}, \mathcal{M}(\textcolor{teal}{\psi^{N_i}_n}))-\mathcal{J}(\textcolor{purple}{\pi_{\theta_{N_i}}}, \mathcal{M}(\textcolor{teal}{\psi^{0}_n}))\Big]
    +\Big[\mathcal{J}(\textcolor{purple}{\pi_{\theta_{0}}}, \mathcal{M}(\textcolor{teal}{\psi^{N_i}_n}))-\mathcal{J}(\textcolor{purple}{\pi_{\theta_{0}}}, \mathcal{M}(\textcolor{teal}{\psi^{0}_n}))\Big]\Bigg]\label{eq:2-2}
\end{align}
\end{figure*}

Fig.~\ref{fig:contribution} shows the normalized amounts of improvements from control-policy optimization (red) and morphology optimization (teal), averaged over 200 initial fall-down poses. We observe that the ratio differs across robots. In Bez3, performance gains mainly stem from control-policy optimization, suggesting that the pretrained policy was suboptimal. By contrast, in OP3-Rot and Sigmaban, morphology contributes most of the improvement, indicating that their initial designs leave significant room for optimization.

\subsection{Necessity of control policy fine-tuning}\label{sec:exp3}
The \ourmethod{} first pretrains a multi-design control policy and then finetunes it using trajectories collected from selected morphologies. An important ablation question is whether this coarse evaluator, the pretrained multi-design policy, already provides sufficient guidance for effective morphology optimization.

Tab.~\ref{tab:prevsfinetune} compares the performance of final morphologies evolved under pretrained and finetuned control policies. Each value is computed as $\frac{\text{pretrained}-\text{base}}{\text{finetuned}-\text{base}}$, with the best optimizer selected for the finetuned policy from Tab. \ref{tab:effectiveness}. The results are consistent with Fig.~\ref{fig:contribution}: the weaker the control policy’s contribution to co-design, the higher the relative performance of morphologies evolved under the pretrained policy. Robot Wolfgang is omitted here, as its co-design process yields no performance improvement.
\begin{table}[!tb]
\caption{Performance of final morphologies evolved under pretrained vs. finetuned control policies.}\label{tab:prevsfinetune}
\centering
\begin{tabular}{cccc}
\bottomrule
                       & Bez1    & Bez2     & Bez3  \\ \hline
Pretrained vs. Finetuned & 0.46    & 0.73     & 0.32  \\ \hline
                       & OP3-Rot & Sigmaban & NUGUS \\ \hline
Pretrained vs. Finetuned & 0.81    & 0.77     & 0.29  \\ \bottomrule
\end{tabular}
\end{table}

\section{Conclusion}
In this paper, we introduce \ourmethod{}, a scalable and efficient framework that co-designs humanoid robots for fall recovery. 
\ourmethod{} progressively uncovers optimized morphologies from initial designs guided by finetuned policies derived from a pretrained multi-design control policy. The framework is compatible with multiple humanoid robot models and optimization algorithms.
Experiments on seven public humanoid robots validate the effectiveness of \ourmethod{}, yielding an average performance improvement of $44.55\%$.
In addition, experiments show that the individual optimization of morphology propels at least $40\%$ of improvements in co-designing four humanoid robots, manifesting the significance of humanoid co-design.

\ourmethod{} is also able to adapt to other tasks, if the environments of other tasks are available.
Future directions include leveraging \ourmethod{} to benchmark across diverse public humanoid robots, extending to multiple tasks, and ensuring compatibility with various optimizers to further advance research on humanoid co-design. Real-world validation of humanoid co-design also presents an important avenue, despite the associated costs.

{\small
\balance
\bibliographystyle{IEEEtran}
\bibliography{references}
}

\end{document}